\documentclass{article} 
\usepackage{iclr2027_conference,times}

\usepackage{amsmath,amsfonts,bm}

\def\eqref#1{equation~\ref{#1}}

\def\1{\bm{1}}

\DeclareMathAlphabet{\mathsfit}{\encodingdefault}{\sfdefault}{m}{sl}
\SetMathAlphabet{\mathsfit}{bold}{\encodingdefault}{\sfdefault}{bx}{n}

\usepackage{hyperref}
\usepackage{url}

\usepackage{graphicx}

\usepackage{amsmath}
\usepackage{amssymb}

\usepackage{booktabs}
\usepackage{multirow}
\usepackage{array}
\usepackage{tabularx}
\usepackage{adjustbox}

\usepackage{algorithm}
\usepackage{algorithmic}

\usepackage{xcolor}
\usepackage{tikz}
\usepackage{pgfplots}
\pgfplotsset{compat=1.18}

\usepackage{float}

\title{GSM: Efficient Language Modeling with Shared Global State}

\author{
Yunao Zheng\textsuperscript{1}
\quad
Bin Wen\textsuperscript{2,*,$\dagger$}
\quad
Xiaojie Wang\textsuperscript{1,$\dagger$}
\\[2pt]
Kaiyu Jiang
\quad
Xuanyu Zheng
\quad
Changyi Liu
\quad
Hongyi Fu
\\
Jianxiong Wang
\quad
Tianke Zhang
\quad
Haonan Fan
\quad
Yingxin Li
\\
Jiankang Chen
\quad
Xu Wang
\quad
Tingting Gao
\quad
Han Li
}

\iclrfinalcopy

\begin{document}

\maketitle

\begingroup
\renewcommand{\thefootnote}{}
\footnotetext{
\textsuperscript{1} Beijing University of Posts and Telecommunications.
\quad
\textsuperscript{2} Kuaishou Technology.
\quad
\textsuperscript{*} Project Leader.
\quad
\textsuperscript{$\dagger$} Corresponding authors: Xiaojie Wang and Bin Wen.
}
\endgroup

\begin{abstract}
Improving language model efficiency requires not only cheaper operations but also eliminating redundant computation. We argue that effective modeling requires information accessibility and computational depth, but not repeated historical retrieval across layers. Guided by this principle, we introduce the Global State Model (GSM), a causal encoder--decoder architecture that separates historical information acquisition from deep computation. The encoder progressively aggregates long-range information through multi-stage sparse retrieval into a fixed-size shared state, while decoder layers perform successive transformations over this state without repeated historical access. This design makes decoder attention and cache costs independent of history length, while dependency-aware pruning further reduces prefill computation. Experiments across multiple model scales show that GSM achieves downstream performance comparable to matched DeepSeek-V4.1-Flash-based baselines, with improvements at some scales, while preserving long-context retrieval capabilities. GSM also improves training and inference throughput and reduces cache overhead. These results demonstrate that eliminating architectural redundancy offers an effective path toward efficient language modeling without sacrificing information accessibility or computational depth.
\end{abstract}

\section{Introduction}
\label{sec:introduction}

Architectural design for efficient language models typically focuses on reducing the cost of individual operations. An equally important question, however, is whether all the computation performed by a model is necessary. In deep models, different layers may repeatedly retrieve and process similar historical information or perform computations that do not contribute to the target representation. Such redundancy accumulates with sequence length and model depth. Improving architectural efficiency therefore requires not only cheaper operations but also identifying and eliminating unnecessary computation.

Eliminating redundancy must not compromise the capabilities required for effective task performance. We consider two essential requirements: \emph{information accessibility} and \emph{computational depth}. The former requires the model to access task-relevant information, including distant historical context, while the latter requires information to be progressively processed and integrated through interdependent transformations across layers. However, information accessibility does not require each layer to independently retrieve historical information, nor does computational depth require every layer to repeat the process of acquiring that information. Architectural design should therefore preserve the necessary information access and deep computation while eliminating redundant operations involved in realizing these capabilities.

Existing efficient architectures have reduced such redundancy to varying degrees. Hybrid attention architectures combine efficient state updates with a small number of global attention layers to reduce explicit historical access~\cite{kimi2026k3}. Sparse attention architectures instead lower the cost of processing historical information through selective retrieval~\cite{deepseek2026v4}. DeepSeek-V4.1-Flash further reduces redundancy through its causal encoder--decoder architecture, cross-layer KV sharing, and partial reuse of sparse retrieval indices, avoiding repeated construction of historical representations and some redundant retrieval operations~\cite{deepseek2026v41,brandon2024crosslayer}. Nevertheless, its decoder layers still access explicit historical KV representations and perform attention aggregation. Repeated access to and aggregation over historical KV representations are not prerequisites for preserving computational depth: information acquisition can be shared, while the processing of that information can proceed through successive layers. This observation suggests an opportunity to further reduce architectural redundancy.

Motivated by this insight, we propose the Global State Model (GSM), which reduces unnecessary computation by reorganizing historical information acquisition and deep computation. GSM adopts a causal encoder--decoder architecture. Through multi-stage sparse retrieval, the encoder combines the current hidden representation with accumulated retrieval outputs to progressively select and aggregate long-range information, integrating it into representations at recent positions to form a fixed-size shared state. Rather than independently retrieving from explicit history, each decoder layer generates queries from the preceding layer's output and performs attention over the same KV representation derived from this shared state. GSM thus concentrates historical information acquisition and aggregation in the encoder while preserving the decoder's ability to perform distinct, interdependent transformations across layers, reducing redundant long-range historical access across model depth.

This design provides additional efficiency benefits. Because the decoder accesses only a fixed-size shared state, its per-step attention cost and corresponding state cache requirements do not grow with history length. During prefill, GSM can further skip encoder computations that cannot affect the target shared state through dependency-aware pruning. Importantly, the fixed-size constraint applies only to the state accessible to the decoder. The encoder retains explicit historical representations, allowing subsequent positions to retrieve information that has not yet been incorporated into the shared state. GSM therefore achieves efficiency not by discarding historical information, but by reorganizing how that information is acquired and used.

We evaluate GSM against matched baselines built on the DeepSeek-V4.1-Flash architecture across multiple parameter scales. GSM achieves comparable performance across downstream tasks, with improvements at some model scales. Long-context experiments further demonstrate its ability to utilize distant historical information. Meanwhile, GSM improves both training and inference throughput while reducing cache requirements. Together, these results demonstrate that eliminating architectural redundancy while preserving information accessibility and computational depth provides an effective approach to improving language model efficiency.

\begin{figure*}[t]
    \centering
    \includegraphics[width=\textwidth]{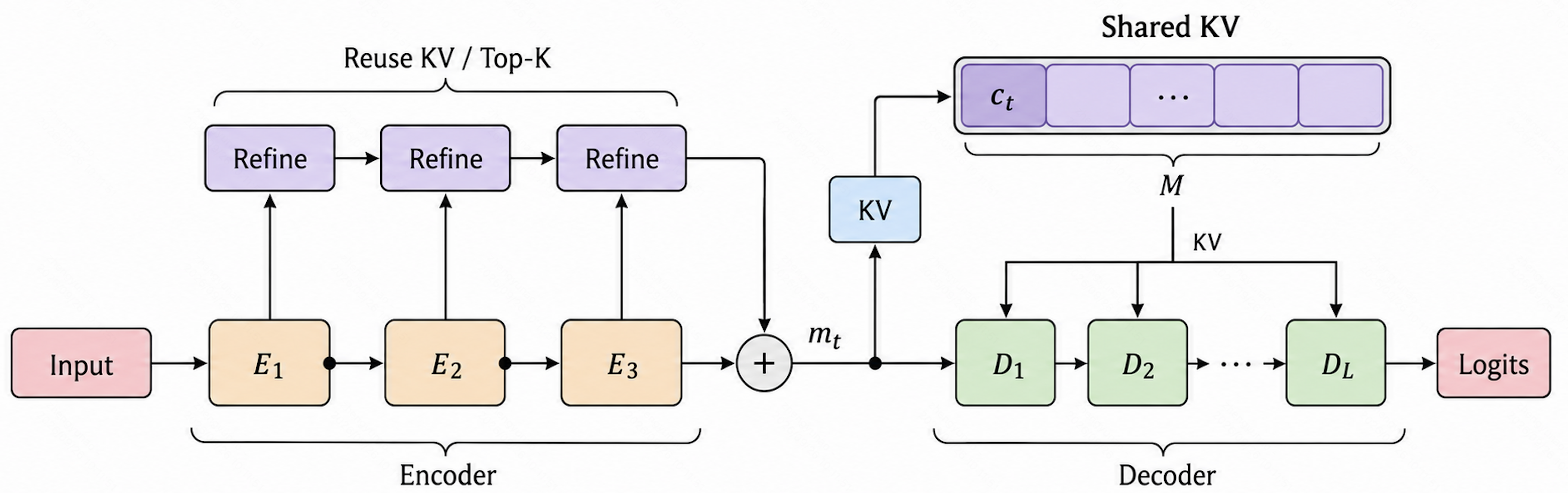}
    \caption{Overview of the GSM architecture. The encoder progressively refines historical information, while the decoder layers share a single fixed-size KV memory.}
    \label{fig:gsm_architecture}
\end{figure*}

\section{Related Work}
\label{sec:related_work}

\subsection{From Shared Historical Representations to Shared States}
\label{sec:related_sparse}

Cross-layer KV sharing reduces the redundancy of maintaining distinct historical representations at every layer. Cross-Layer Attention (CLA) shares keys and values across adjacent layers~\cite{brandon2024crosslayer}, while Multi-Layer Key-Value (MLKV) sharing extends this principle across both attention heads and layers~\cite{zuhri2024mlkv}. Layer-Condensed KV Cache (LCKV) further reduces the number of layers that explicitly compute and retain KV representations~\cite{wu2024layercondensed}. You Only Cache Once (YOCO) adopts a decoder--decoder architecture in which a causal self-decoder constructs a global KV cache that is subsequently accessed by cross-decoder layers using layer-specific queries~\cite{sun2024yoco}. Collectively, these approaches demonstrate that layer-specific computation does not require layer-specific historical KV caches.

Sharing historical representations, however, does not eliminate the cost of history-dependent retrieval. In YOCO, each cross-decoder layer still performs global attention over a shared cache whose size grows with context length. You Only Index Once introduces Cross-Layer Sparse Attention (CLSA), which performs token-level Top-$K$ selection once and reuses both the KV cache and selected indices across cross-decoder layers~\cite{sun2026clsa}. DeepSeek-V4.1-Flash combines shared historical KV with sparse attention, local windows, and selective reindexing, allowing selected decoder layers to revise their retrieval results within a shared candidate pool~\cite{deepseek2026v41}. These methods progressively reduce redundant KV storage and repeated routing, but the decoder still retrieves from a history-length-dependent KV bank.

GSM extends this progression from sharing historical representations and retrieval indices to sharing aggregated historical information. Its encoder performs multi-stage historical retrieval and progressively integrates the accumulated readout into recent representations, from which a fixed-size shared KV state is constructed. Each decoder layer then queries the same state using the output of the preceding layer, preserving layer-wise computational depth without decoder-side long-range indexing. Consequently, the shared object in GSM is no longer a KV bank spanning the explicit history, but a bounded state that jointly represents recent context and retrieved long-range information.

To characterize the resulting computational difference, let $N$ denote the history length, $L_D$ the number of decoder layers, $K$ the sparse retrieval budget, $W$ the local window size, $P$ the candidate-pool capacity, $J$ the number of Reindex layers, and $M$ the GSM state size. Let $d_I$ and $d$ denote the effective widths of index scoring and main attention, respectively. In DeepSeek-V4.1-Flash, the first Full layer indexes the historical KV bank, Reindex layers rescore the candidate pool, and Reuse layers retain the previous selections. Omitting token-wise projections and feed-forward computation, the per-token decoder attention costs are
\begin{align}
T_{\mathrm{DS},D}^{\mathrm{decode}}
&=
\mathcal{O}\!\left(
Nd_I + JPd_I + L_D(K+W)d
\right), \\
T_{\mathrm{GSM},D}^{\mathrm{decode}}
&=
\mathcal{O}\!\left(
L_DMd
\right).
\end{align}
GSM therefore removes history-dependent indexing from the decoder and bounds each decoder layer's attention to $M$ shared state entries.

Let $d_{\mathrm{KV}}$ denote the effective storage width of a KV entry. When local-window caches are retained during autoregressive decoding, the decoder-accessible KV storage scales as
\begin{align}
C_{\mathrm{DS},D}
&=
\mathcal{O}\!\left(
Nd_{\mathrm{KV}} + L_DWd_{\mathrm{KV}}
\right), \\
C_{\mathrm{GSM},D}
&=
\mathcal{O}\!\left(
Md_{\mathrm{KV}}
\right).
\end{align}
Thus, both the attention cost and decoder-accessible state size of GSM are independent of history length. GSM does not discard the explicit history: the encoder retains a historical cache for subsequent retrieval and state construction, while the decoder operates exclusively on the bounded shared state.

\subsection{Connections to Linear Attention}
\label{sec:related_linear}

Linear attention summarizes historical information in a fixed-size state~\cite{katharopoulos2020transformers}. For example, the update in DeltaNet can be abstracted as~\cite{schlag2021linear,yang2024parallelizing}
\begin{equation}
S_t^{\mathrm{lin}}
=
A(x_t)S_{t-1}^{\mathrm{lin}}
+
\beta_t k_t v_t^{\top},
\label{eq:linear_state}
\end{equation}
where newly written information is generated primarily from the input at the current position. Such linear attention methods can use historical information through a recurrent state. Once the original history has been compressed, however, subsequent positions can no longer directly retrieve individual entries from it.

GSM differs in that the process of writing to the shared state retains access to an explicit representation of the history. At each stage, it can generate a new query from the current hidden representation and the accumulated retrieval output, and access the full causal history again. Subsequent retrieval can thus depend on information obtained earlier, rather than being determined solely by the input at the current position. Its temporal update can therefore be abstracted as
\begin{equation}
S_t
=
A S_{t-1}
+
\mathcal{B}_t
\left(
x_t;K_{\leq t},V_{\leq t}
\right).
\label{eq:gsm_state}
\end{equation}
The main distinction between GSM and linear attention is therefore not whether they use a previous state, but whether constructing a new state can involve revisiting the explicit history. Linear attention can only further transform the compressed state, whereas GSM can retrieve historical information again based on new intermediate representations. Accordingly, GSM must retain an encoder-side history cache: it is the decoder state size, rather than the total historical storage of the full model, that remains fixed.

\subsection{Multi-Step State Updates and Chain-of-Thought}
\label{sec:related_cot}

DeltaNet applies one rank-one state correction per input, with the expressiveness of each update constrained by its form. DeltaProduct, proposed by Siems et al., extends this process to multiple updates to increase the expressiveness of state transitions~\cite{siems2025deltaproduct}. Experiments show improvements in state tracking, language modeling, and some in-context retrieval tasks, as well as substantially stronger length extrapolation. However, the number of updates is specified in advance, making it difficult to allocate computation dynamically according to the demands of different inputs. Increasing the number of updates also incurs substantial computational overhead. These limitations have hindered the broader adoption of multi-step state updates.

GSM faces a similar information bottleneck: although encoding involves multiple rounds of history retrieval, the new information at each position must ultimately be incorporated into a finite-dimensional hidden representation. This raises two questions: whether a single write can accommodate the information required by the current computation, and how to supplement the shared state when subsequent decoder computation creates new information needs. Chain-of-thought (CoT) offers a way to address this bottleneck by generating intermediate reasoning steps~\cite{wei2022chainofthought}. In GSM, each generated token passes through the encoder at the next step, providing another opportunity to retrieve historical information and write to the state. If the decoder expresses intermediate conclusions or new information needs in subsequently generated tokens, the encoder can adjust retrieval accordingly and incorporate additional information into the shared state. CoT can therefore distribute information acquisition and processing across multiple time steps, allowing the shared state to be continually updated around the current reasoning step without requiring a single write to anticipate all subsequent information needs.

DeepSeek-V4.1-Flash has the same state capacity as GSM in this comparison, since attention in both models operates over the same number of hidden representations. The difference is that DeepSeek-V4.1-Flash retains KV covering the history and allows some decoder layers to reselect entries using new queries. By contrast, the GSM decoder accesses only a fixed window and relies on subsequent encoding steps to incorporate additional historical information. Thus, once CoT addresses the information bottleneck described above, GSM should, in principle, be capable of matching the strong long-context processing capabilities of leading models.

\section{Model Architecture}
\label{sec:architecture}

As illustrated in Figure~\ref{fig:gsm_architecture}, GSM adopts a causal encoder--decoder architecture that separates historical information aggregation from layer-by-layer decoding. The encoder aggregates historical information through multiple retrieval stages and fuses the accumulated output with its final encoded representation to construct a shared KV state with a fixed window size. Each decoder layer accesses the same state using queries updated from the preceding layer, preserving computational dependencies across layers while avoiding repeated construction of historical KV and long-range indexing.

\subsection{Multi-Stage History Aggregation}
\label{sec:architecture_aggregation}

Let $d$ denote the model's hidden dimension and $S$ the number of aggregation stages. Let $\mathbf{h}_t^{(s)}\in\mathbb{R}^{d}$ denote the encoder representation at position $t$ in aggregation stage $s$, obtained by mixing the multi-stream residual representations of the corresponding layer~\cite{xie2025mhc}. Let $\mathbf{r}_t^{(s)}$ denote the accumulated retrieval output through that stage, initialized as $\mathbf{r}_t^{(0)}=\mathbf{0}$. The query input at each stage is
\begin{equation}
\mathbf{u}_t^{(s)}
=
\begin{cases}
\mathbf{h}_t^{(1)}, & s=1,\\
[\mathbf{h}_t^{(s)};\mathbf{r}_t^{(s-1)}], & s>1,
\end{cases}
\end{equation}
where $[\cdot;\cdot]$ denotes concatenation along the feature dimension. Queries are generated using stage-specific root mean square normalization (RMSNorm)~\cite{zhang2019rmsnorm} and query projections:
\begin{equation}
\mathbf{q}_t^{(s)}
=
\operatorname{Query}_s
\left(
\operatorname{RMSNorm}_s(\mathbf{u}_t^{(s)}),p_t
\right),
\end{equation}
where $p_t$ denotes the position metadata required for positional encoding. The input dimension is $d$ in the first stage and $2d$ in subsequent stages, allowing later queries to depend on both the current encoded representation and previously retrieved information.

The aggregation branch directly reuses the local window $\mathcal{W}_t^{(s)}$, historical KV bank $\mathcal{B}^{(s)}$, and Top-$K$ indices $\mathcal{I}_t^{(s)}$ used by the corresponding encoder layer, without a separate indexer. Retrieval and accumulation proceed as follows:
\begin{align}
\Delta\mathbf{r}_t^{(s)}
&=
\operatorname{Output}_s\!\left(
\operatorname{Attn}\!\left(
\mathbf{q}_t^{(s)};
\mathcal{W}_t^{(s)},
\mathcal{B}^{(s)}[\mathcal{I}_t^{(s)}]
\right)
\right),\\
\mathbf{r}_t^{(s)}
&=
\mathbf{r}_t^{(s-1)}+\Delta\mathbf{r}_t^{(s)}.
\end{align}
Here, the aggregation query determines the attention weights over the selected entries; it does not reselect the Top-$K$ entries. Multi-stage aggregation therefore progressively adjusts the retrieved information by updating queries while reusing the selections already made by the encoder.

The accumulated retrieval output is updated across encoding stages without being written back to the main encoder branch between stages. The encoder thus retains its original computational path across layers, and the aggregation result is fused with its output only after encoding is complete.

\subsection{Shared State Construction}
\label{sec:architecture_memory}

Let $\mathbf{e}_t$ denote the mixture of the final multi-stream encoder representations. The model fuses it with the accumulated retrieval output and applies a shared KV mapping to generate a state entry:
\begin{align}
\mathbf{z}_t
&=
\mathbf{e}_t+\mathbf{r}_t^{(S)},\\
\mathbf{m}_t
&=
\operatorname{KV}_{\mathrm{sh}}(\mathbf{z}_t,p_t).
\end{align}
This mapping applies input normalization, KV projection, KV normalization, and positional encoding in sequence, producing one state entry per position without compression along the sequence dimension.

At position $t$, the decoder accesses only the shared KV entries corresponding to the most recent $M$ valid positions:
\begin{equation}
\mathcal{M}_t
=
\left[
\mathbf{m}_{\max(1,t-M+1)},\ldots,\mathbf{m}_t
\right],
\qquad M=W+K,
\end{equation}
where $W$ and $K$ denote the encoder's local window size and number of sparsely retrieved entries, respectively. The actual accessible range is constrained by document boundaries and position masks. The decoder's retrieval budget therefore matches that of a single encoder attention operation. This window limits only the decoder's direct access: distant information has already been aggregated into the representations within the window by the encoder, allowing the shared state to carry both recent and long-range information.

The fixed-size shared window also allows encoder computation during prefill to be pruned according to its dependencies. For an input of length $N$ that requires only next-token prediction, only the shared KV entries for the final $m=\min(N,M)$ positions need to be constructed. It is therefore unnecessary to process every position through every encoder layer. Let $L_E$ denote the number of encoder layers and $\ell_\star$ the last historical KV source layer, with layers indexed from $1$. Once full-sequence KV construction at this layer is complete, each layer $\ell\geq\ell_\star$ performs attention retrieval and subsequent within-block computation only for the final
\begin{equation}
R_\ell
=
\min\!\left(
N,\,
m+(L_E-\ell)(W-1)
\right)
\end{equation}
positions. The additional positions cover the dependencies introduced by subsequent local windows, allowing the computation range to shrink progressively with depth until only the final $m$ positions remain. Each aggregation stage likewise needs to compute retrieval outputs only for these positions. Finally, the model constructs the shared state from the fused representations at the final $m$ positions and initializes the decoder with the representation at the last position, reducing prefill computation in the later encoder layers while preserving all computational dependencies.

\subsection{Shared State Decoding}
\label{sec:architecture_decoder}

The decoder initializes the representation at the current position with $\mathbf{z}_t$. The attention module in each layer retains only its own query projection, output projection, and attention sink; it no longer generates independent KV or performs long-range indexing. Let $\mathbf{d}_t^{(\ell-1)}$ denote the mixed representation entering the attention branch of layer $\ell$. Then,
\begin{align}
\mathbf{q}_{t,D}^{(\ell)}
&=
\operatorname{Query}_{D,\ell}
\left(
\operatorname{RMSNorm}_{D,\ell}
(\mathbf{d}_t^{(\ell-1)}),p_t
\right),\\
\mathbf{a}_{t,D}^{(\ell)}
&=
\operatorname{Output}_{D,\ell}
\left(
\operatorname{Attn}
(\mathbf{q}_{t,D}^{(\ell)};\mathcal{M}_t)
\right).
\end{align}
All decoder layers at the same position access the same $\mathcal{M}_t$, but their queries change as representations are updated across layers. Sharing KV across layers therefore preserves sequential computational dependencies across depth. The shared state remains unchanged within the decoder, and no additional history retrieval is performed.

\section{Experiments}
\label{sec:experiments}

We evaluate whether GSM can reduce the cost of accessing historical information without substantially compromising model performance. We compare GSM with a baseline based on the DeepSeek-V4.1-Flash architecture at matched model scales and training budgets. The baseline already combines sparse attention, KV compression, and cross-layer sharing, allowing us to assess the additional benefits of GSM over an efficient attention architecture. Detailed experimental settings are provided in Appendix~\ref{app:experiments}.

\subsection{Downstream Task Performance}
\label{sec:exp_downstream}

We evaluate a 1.5B dense model and 8B-A1.5B and 30B-A3B mixture-of-experts (MoE) models on ten downstream tasks covering commonsense reasoning, language understanding, and knowledge-based question answering. For each scale, we evaluate the checkpoint obtained at the end of the main training phase. Avg.\ denotes the unweighted mean across the ten tasks. We report single-run results for each configuration; variability across random seeds is not assessed.

\begin{table*}[t]
\centering
\small
\setlength{\tabcolsep}{8pt}
\caption{Downstream task performance (\%). Metrics for tasks marked with $\dagger$ are specified in Appendix~\ref{app:quality_protocol}.}
\label{tab:downstream}
\begin{tabular}{lrrrrrr}
\toprule
\multirow{2}{*}{Task} & \multicolumn{2}{c}{1.5B} & \multicolumn{2}{c}{8B-A1.5B} & \multicolumn{2}{c}{30B-A3B} \\
\cmidrule(lr){2-3}\cmidrule(lr){4-5}\cmidrule(lr){6-7}
& Baseline & GSM & Baseline & GSM & Baseline & GSM \\
\midrule
ARC-Easy$^{\dagger}$~\cite{clark2018arc} & \textbf{59.09} & 58.59 & \textbf{72.90} & 71.55 & \textbf{75.13} & 74.79 \\
ARC-Challenge$^{\dagger}$ & \textbf{33.45} & 33.36 & \textbf{36.69} & \textbf{36.69} & 39.93 & \textbf{41.81} \\
PIQA$^{\dagger}$~\cite{bisk2020piqa} & \textbf{70.24} & 69.80 & 73.23 & \textbf{74.43} & \textbf{75.95} & 75.30 \\
SciQ$^{\dagger}$~\cite{welbl2017sciq} & 81.90 & \textbf{82.40} & \textbf{89.80} & \textbf{89.80} & 91.80 & \textbf{92.70} \\
BLiMP~\cite{warstadt2020blimp} & 80.18 & \textbf{80.79} & 81.66 & \textbf{82.24} & \textbf{82.25} & 81.68 \\
HellaSwag$^{\dagger}$~\cite{zellers2019hellaswag} & \textbf{50.13} & 49.35 & 44.52 & \textbf{44.65} & 48.91 & \textbf{50.09} \\
WinoGrande~\cite{sakaguchi2020winogrande} & \textbf{54.62} & 54.14 & 55.41 & \textbf{57.14} & \textbf{59.27} & 58.25 \\
MMLU~\cite{hendrycks2021mmlu} & 25.47 & \textbf{26.41} & 25.75 & \textbf{27.10} & 26.57 & \textbf{26.82} \\
OpenBookQA$^{\dagger}$~\cite{mihaylov2018openbookqa} & \textbf{37.60} & 36.60 & \textbf{29.00} & 28.80 & 28.60 & \textbf{31.00} \\
LAMBADA~\cite{paperno2016lambada} & \textbf{41.84} & 41.12 & \textbf{46.79} & 46.30 & \textbf{52.53} & 52.32 \\
\midrule
Avg. & \textbf{53.45} & 53.26 & 55.58 & \textbf{55.87} & 58.09 & \textbf{58.48} \\
\bottomrule
\end{tabular}
\end{table*}

Table~\ref{tab:downstream} shows that GSM achieves average performance comparable to the baseline at the dense model scale and slightly higher average scores in both MoE configurations. These results suggest that the encoder-aggregated shared state can support subsequent layer-by-layer computation without a substantial loss of downstream performance. The comparable results across the tested dense and MoE configurations further suggest that this behavior is not limited to a particular model scale or feed-forward network (FFN) architecture.

\subsection{Long-Context Modeling}
\label{sec:exp_long_context}

We evaluate GSM's ability to retrieve and integrate information from long contexts using RULER~\cite{hsieh2024ruler}. We first evaluate a 1.5B GSM on three retrieval tasks across context lengths from 4K to 64K, comparing direct-answer and short-CoT variants after task adaptation. Detailed settings are provided in Appendix~\ref{app:quality_protocol}.

\begin{table}[t]
\centering
\footnotesize
\setlength{\tabcolsep}{2.5pt}
\caption{RULER retrieval accuracy of 1.5B GSM (\%).}
\label{tab:ruler}
\begin{tabular}{lrrrrrr}
\toprule
Answer Format & 4K & 8K & 16K & 32K & 64K & Avg. \\
\midrule
Direct Answer & 98.33 & 100.00 & 98.33 & 100.00
              & 98.33 & 99.00 \\
Short CoT     & 100.00 & 100.00 & 100.00 & 100.00
              & 100.00 & 100.00 \\
\bottomrule
\end{tabular}
\end{table}

As shown in Table~\ref{tab:ruler}, GSM maintains near-perfect retrieval accuracy across 4K--64K contexts, with short CoT achieving 100\% accuracy at every evaluated length. These results demonstrate effective long-range retrieval despite the decoder's fixed-size shared state, while suggesting that intermediate generation can further support information acquisition.

We further investigate million-token information utilization using the 30B-A3B GSM. After YaRN~\cite{peng2024yarn} extension and supervised adaptation on RULER-derived tasks, we evaluate multi-key retrieval and multi-hop reasoning on prompts containing nearly 1M tokens. Relevant facts are distributed throughout the context, requiring the model to retrieve multiple targets or follow multi-hop dependencies. The encoder retains sparse access to historical information, while each decoder layer attends to only $M=640$ shared state positions ($W=128$, $K=512$).

\begin{table}[t]
\centering
\footnotesize
\setlength{\tabcolsep}{6pt}
\caption{1M RULER-derived accuracy of 30B-A3B GSM.}
\label{tab:ruler_1m}
\begin{tabular}{lrr}
\toprule
Task & Correct / Total & Accuracy (\%) \\
\midrule
Multi-key Retrieval & 64 / 64 & 100.00 \\
Multi-hop Reasoning & 59 / 64 & 92.19 \\
\midrule
Overall & 123 / 128 & 96.09 \\
\bottomrule
\end{tabular}
\end{table}

Table~\ref{tab:ruler_1m} shows that GSM achieves 100\% accuracy on multi-key retrieval and 92.19\% on multi-hop reasoning, including 86.36\% accuracy on eight-hop problems. These results demonstrate that GSM can retrieve and integrate information across nearly 1M tokens while restricting decoder attention to only 640 shared state positions. 

Together, these evaluations demonstrate that GSM supports effective long-context modeling, extending to one million tokens.

\subsection{Computational Efficiency}
\label{sec:exp_efficiency}

We evaluate the training and inference efficiency of the 30B-A3B models on input lengths from 4K to 256K using checkpoints after 64K YaRN adaptation. Training is benchmarked on 32 GPUs (16 H800 and 16 H100), while inference is measured separately on individual H800 and H100 GPUs with batch size 1. Detailed settings are provided in Appendix~\ref{app:efficiency_protocol}.

\begin{figure*}[t]
\centering
\includegraphics[width=0.8\textwidth]{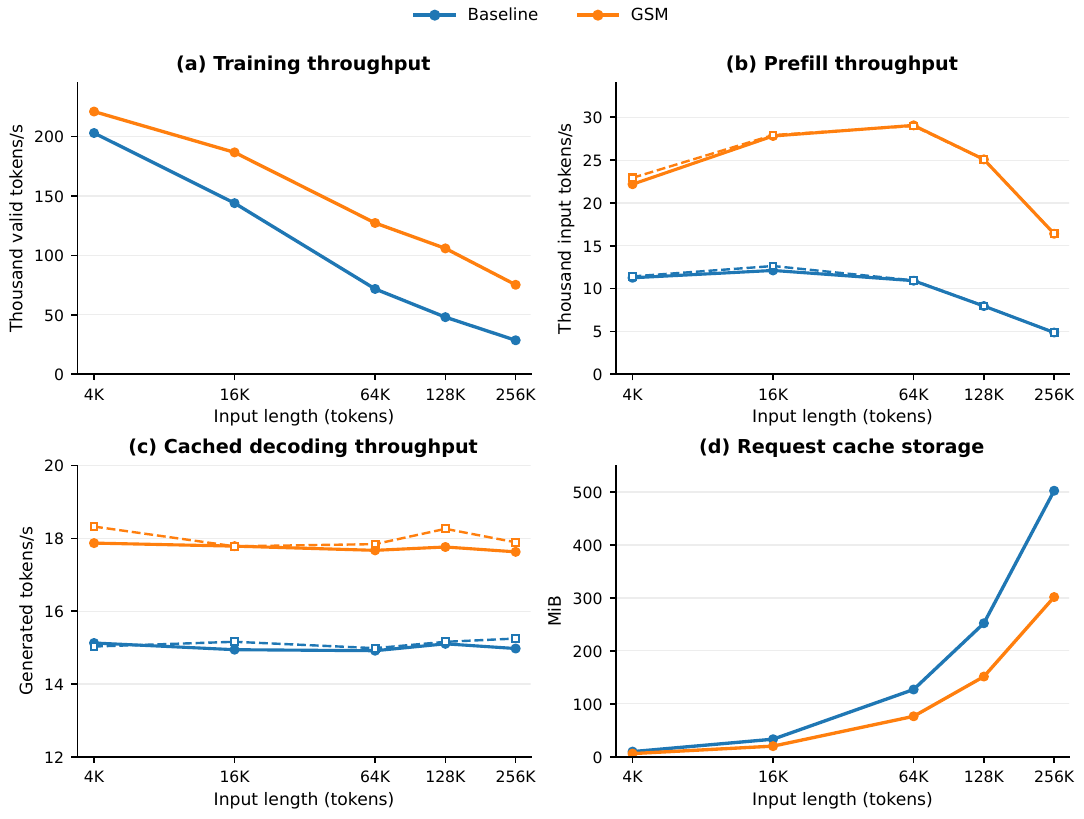}
\caption{Efficiency comparison of the 30B-A3B models across input lengths of 4K--256K. (a) Training throughput on 32 GPUs. (b,c) Prefill and cached decoding throughput on H800 (solid lines) and H100 (dashed lines), with batch size 1 and 64 decoding steps. (d) Request-cache storage after prefill.}
\label{fig:inference_efficiency}
\end{figure*}

Figure~\ref{fig:inference_efficiency}(a) shows that GSM improves training throughput across all evaluated lengths, with the speedup increasing from 1.09$\times$ at 4K to 1.77$\times$ at 64K and 2.63$\times$ at 256K. At 256K, GSM reduces the training step time from 36.70 to 13.93 seconds. The increasing advantage demonstrates that consolidating long-range information processing in the encoder reduces repeated computation over long histories. GSM has higher peak training memory, with allocations of 56.75 GiB per GPU compared with 50.61 GiB for the baseline at 256K.

GSM also achieves substantially higher prefill throughput (Figure~\ref{fig:inference_efficiency}(b)), with speedups increasing from approximately 2.0$\times$ at 4K to 3.37$\times$ at 256K on both GPU types. Cached decoding is consistently faster by approximately 1.17--1.22$\times$, with throughput remaining stable as the input length increases (Figure~\ref{fig:inference_efficiency}(c)). By performing long-range retrieval and aggregation in the encoder, GSM enables decoder layers to reuse a fixed-window shared state rather than repeatedly selecting and accessing historical KV representations, improving both prefill and autoregressive decoding efficiency.

As shown in Figure~\ref{fig:inference_efficiency}(d), GSM reduces request-cache storage by approximately 40\% at 256K, from 0.491 to 0.295 GiB. Total peak GPU allocation remains comparable (72.52 GiB for GSM versus 72.80 GiB for the baseline), as model weights and execution buffers dominate memory usage. Overall, these results demonstrate that GSM substantially improves long-context training and generation efficiency at the 30B-A3B scale, with increasing advantages in training and prefill as context length grows, consistently faster decoding, and lower request-cache overhead.

\subsection{Ablation Studies}
\label{sec:exp_ablation}

\subsubsection{Shared History Retrieval}

Using 0.5B models, we compare a separate aggregation indexer with reuse of the main attention Top-$K$ selections. Both variants use the same training budget, accumulated retrieval output, and query formulation, isolating the effect of independent versus shared history selection for the aggregation branch.

\begin{table}[t]
\centering
\footnotesize
\setlength{\tabcolsep}{2.8pt}
\caption{Shared retrieval ablation (0.5B, \%).}
\label{tab:retrieval_ablation}
\begin{tabular}{lrrr}
\toprule
Retrieval Method & Standard Avg. & RULER Avg. & RULER 64K \\
\midrule
Separate Aggregation Indexer & \textbf{64.918} & \textbf{96.67} & \textbf{93.33} \\
Shared Main Attention Top-$K$ & 64.765 & 94.33 & \textbf{93.33} \\
\bottomrule
\end{tabular}
\end{table}

Table~\ref{tab:retrieval_ablation} shows that reusing the main attention Top-$K$ selections changes the standard-task average only slightly (64.918 versus 64.765), while mean RULER accuracy decreases from 96.67\% to 94.33\%. Accuracy at 64K remains identical at 93.33\%. Appendix~\ref{app:retrieval_ablation} shows that the overlap between the two index sets decreases markedly as context length grows, indicating that sharing changes the historical positions accessed by the aggregation branch. Nevertheless, the comparable downstream performance and identical 64K retrieval accuracy suggest that aggregation can use the candidates selected by main attention without a separate history index under the tested conditions.

\subsubsection{Latent Refinement Frequency}
\label{sec:exp_refinement}

We compare three refinement schedules on the 0.5B model: a single refinement at encoder layer 12, the default three refinements at the ends of the KV-sharing groups (layers 6, 10, and 12), and ten refinements across layers 3--12. All variants are trained from scratch on 10B tokens with the same backbone, shared Top-$K$, state window, and training configuration; parameter counts vary with the number of refinement stages. We report average performance over the ten downstream tasks (Avg., \%), held-out perplexity (PPL), PG19 perplexity at 64K, and mean RULER retrieval accuracy across 4K--64K (\%).

\begin{table}[t]
\centering
\footnotesize
\setlength{\tabcolsep}{2.5pt}
\caption{Latent refinement frequency ablation (0.5B).}
\label{tab:refinement_ablation}
\begin{tabular}{rrrrrr}
\toprule
Stages & Params.\ (M) & Avg.$\uparrow$ & PPL$\downarrow$ & PG19$\downarrow$ & RULER$\uparrow$ \\
\midrule
1  & 507.10 & \textbf{48.27} & 14.7383 & 21.4202 & 83.33 \\
3  & 512.35 & 48.17 & 14.6728 & \textbf{21.2836} & \textbf{94.33} \\
10 & 530.71 & 47.64 & \textbf{14.6598} & 21.3452 & \textbf{94.33}    \\
\bottomrule
\end{tabular}
\end{table}

Table~\ref{tab:refinement_ablation} shows that using a single refinement leaves average downstream performance nearly unchanged (48.27 versus 48.17 for the three-stage default) but increases perplexity and reduces mean RULER accuracy from 94.33\% to 83.33\%. Although the single-stage variant improves training and decoding speed, it weakens long-context information use. Increasing the number of refinements to ten provides only a marginal reduction in held-out PPL (14.6598 versus 14.6728), with no gain in mean RULER accuracy and a lower downstream average, despite adding parameters and refinement computation.

These results favor one refinement per KV-sharing group. Successive groups provide distinct historical representations for aggregation, whereas additional refinements within a group revisit the same KV representation with updated queries. Under the tested configuration, these extra reads provide limited benefits, while collapsing aggregation to a single stage compromises long-range retrieval. The default three-stage schedule therefore balances multistage information aggregation against repeated processing of shared KV.

\subsubsection{Complementary Attention in Latent Aggregation}
\label{sec:exp_attention_analysis}

We compare the attention distributions of the main encoder and latent aggregation branches using a 30B-A3B GSM checkpoint on 16 retrieval examples with approximately 1M-token contexts. Both branches share the same Top-$K$ candidates but generate attention weights from different queries.

\begin{table}[t]
\centering
\footnotesize
\setlength{\tabcolsep}{4pt}
\caption{Attention behavior of main attention and latent aggregation. Historical mass is reported as main attention $\rightarrow$ latent aggregation. TV compares head-averaged distributions excluding the sink.}
\label{tab:latent_attention}
\begin{tabular}{lccc}
\toprule
Stage & TV & Top-8 Overlap (\%) & Historical Mass (\%) \\
\midrule
1 & 0.417 & 24.74 & $38.54 \rightarrow 54.34$ \\
2 & 0.342 & 69.81 & $38.10 \rightarrow 65.34$ \\
3 & 0.224 & 85.58 & $45.24 \rightarrow 43.28$ \\
\bottomrule
\end{tabular}
\end{table}

Table~\ref{tab:latent_attention} reveals complementary attention behavior across refinement stages. The first stage exhibits the largest distributional difference, increasing attention to historical blocks containing relevant facts from 4.08\% to 14.08\%. The second stage further emphasizes compressed history, while the third becomes more closely aligned with main attention. These results show that latent aggregation complements main attention by progressively reweighting shared retrieval candidates, rather than simply repeating the same historical information processing.

\subsubsection{The Role of CoT in Long-Context Information Use}
\label{sec:exp_cot}

Starting from the same 1.5B GSM checkpoint adapted to 64K with YaRN, we train direct-answer and short-CoT variants on the same retrieval examples for the same number of training steps. We evaluate only the final answers. Detailed settings are provided in Appendix~\ref{app:cot}.

As shown in Table~\ref{tab:ruler}, short CoT achieves 100\% accuracy at every tested length from 4K to 64K, including 64K inputs beyond the lengths used for task adaptation. Per-task results indicate that the remaining direct-answer errors arise primarily from multi-key retrieval, which short CoT resolves in this evaluation. This pattern is consistent with the analysis in Section~\ref{sec:related_cot}: generating intermediate tokens produces new model states and additional computational steps for history retrieval and state updates, allowing retrieval, verification, and answering to be distributed across successive steps. These findings suggest that multi-step generation can further improve GSM's use of long-range information in the shared state.

\section{Conclusion}
\label{sec:conclusion}

We introduced the Global State Model (GSM), an approach to improving language model efficiency by eliminating architectural redundancy rather than merely reducing the cost of individual operations. Guided by the requirements of information accessibility and computational depth, GSM separates historical information acquisition from deep computation. A causal encoder progressively aggregates long-range information into a fixed-size shared state, allowing decoder layers to perform successive transformations without repeated historical access. This design reduces decoder attention and cache costs, while dependency-aware pruning further eliminates unnecessary computation during prefill. Experiments across multiple model scales show that GSM achieves downstream performance comparable to matched DeepSeek-V4.1-Flash-based baselines, with improvements at some scales, while preserving long-context retrieval capabilities and improving training and inference throughput. These results demonstrate that reducing architectural redundancy while preserving essential modeling capabilities provides an effective direction for efficient language model design.

\bibliographystyle{iclr2027_conference}
\bibliography{refs}

\newpage
\appendix

\section{Experimental Settings and Supplementary Results}
\label{app:experiments}

\subsection{Models and Training}
\label{app:model_training}

The baseline retains sparse and compressed attention, sliding windows, cross-layer sharing, and manifold-constrained hyper-connections (mHC) from the DeepSeek-V4.1-Flash text backbone. At each model scale, the baseline and GSM use matched network depths, hidden dimensions, training data, and optimization settings. GSM uses a fixed shared state window of $M=W+K$, with $M=256$ for the 0.5B, 1.5B, and 8B-A1.5B models and $M=640$ for the 30B-A3B models. The main training phase uses a sequence length of 4,096 and a global batch size of 128. Training budgets are measured in input tokens, with configurations summarized in Table~\ref{tab:training_setup}.

\begin{table}[t]
\centering
\small
\setlength{\tabcolsep}{5pt}
\caption{Main training configurations.}
\label{tab:training_setup}
\begin{tabular}{lrrr}
\toprule
Scale & Layers & FFN & Input Tokens \\
\midrule
0.5B & 24 & Dense & 10B \\
1.5B & 28 & Dense & 15B \\
8B-A1.5B & 28 & MoE & 18B \\
30B-A3B & 28 & MoE & 31B \\
\bottomrule
\end{tabular}
\end{table}

The 0.5B and 1.5B models have hidden dimensions of 1,024 and 2,048, respectively, and FFN intermediate dimensions of 4,608 and 6,144. Half of the layers are assigned to the encoder and half to the decoder. Aggregation stages are placed at layers 6, 10, and 12 in the 0.5B model and at layers 6, 10, and 14 in the 1.5B model. Both dense models use an encoder local window size of 128 and a sparse retrieval budget of 128 entries. For the 8B-A1.5B configuration, the baseline and GSM have 8.001B and 8.033B total parameters, respectively, with 1.497B and 1.529B active parameters. The 30B-A3B configuration has a hidden dimension of 1,536, an encoder local window size of 128, and a sparse retrieval budget of 512 entries. Its baseline and GSM have 30.640B and 30.659B total parameters, respectively, including a shared vision encoder and projector, with 3.025B and 3.044B active text parameters. Both are jointly trained from scratch on matched text--image data with a 15\% multimodal input quota; the baseline uses the adapted DeepSeek-V4.1-Flash backbone rather than the released model. The 0.5B configuration is retained for the ablation studies.

\subsection{Task Evaluation and Long-Context Settings}
\label{app:quality_protocol}

\textbf{Downstream Evaluation.}

Downstream tasks are evaluated using lm-evaluation-harness. MMLU uses a 5-shot setting, while all other tasks use a 0-shot setting. The maximum context length is 4,096, and the evaluation seed is 42. Tasks marked with $\dagger$ in Table~\ref{tab:downstream} use \texttt{acc\_norm} for dense models and \texttt{acc} for MoE models; all remaining tasks use \texttt{acc}. For BLiMP and MMLU, subtask scores are first aggregated by sample count, after which the resulting task scores are averaged equally with those of the other tasks.

\textbf{1.5B RULER Evaluation.}

We evaluate three retrieval tasks: \texttt{niah\_multikey\_1}, \texttt{niah\_single\_1}, and \texttt{niah\_single\_2}. Each task contains 20 examples at each context length of 4K, 8K, 16K, 32K, and 64K, yielding 300 examples. Direct-answer and short-CoT variants use identical test sets and are evaluated solely on their final answers. Table~\ref{tab:ruler} reports average accuracy across tasks at each length, while Table~\ref{tab:ruler_tasks} provides per-task results.

\begin{table}[t]
\centering
\footnotesize
\setlength{\tabcolsep}{5pt}
\caption{Per-task RULER accuracy of 1.5B GSM (\%).}
\label{tab:ruler_tasks}
\begin{tabular}{lrr}
\toprule
Task & Direct Answer & Short CoT \\
\midrule
\texttt{niah\_multikey\_1} & 97.00 & 100.00 \\
\texttt{niah\_single\_1} & 100.00 & 100.00 \\
\texttt{niah\_single\_2} & 100.00 & 100.00 \\
\bottomrule
\end{tabular}
\end{table}

\textbf{30B-A3B Million-Token Evaluation.}

We extend the 30B-A3B GSM to 1,048,576 tokens using YaRN with a scaling factor of 256, starting from its pretrained and 64K-adapted checkpoint. The original GSM architecture is preserved, with an encoder local window of 128, a sparse retrieval budget of 512, and a decoder shared state size of 640. The model undergoes 4,768 updates of supervised task adaptation, processing approximately 5B input tokens and 0.91M supervised CoT and answer tokens.

We construct two RULER-derived tasks with relevant facts distributed across nearly 1M-token contexts. Multi-key retrieval requires identifying four or eight queried values among 64 random key--value pairs. Multi-hop reasoning requires following variable-assignment chains of two, four, or eight hops, with multiple chains interleaved throughout the context. Both tasks use automatically generated CoT supervision and final answers.

Evaluation uses 128 examples, including 64 multi-key retrieval and 64 multi-hop reasoning cases. All prompts contain 1,047,849--1,047,872 tokens and are processed without history truncation or external retrieval. Greedy decoding is used, and accuracy is measured by exact matching of the complete final answer, requiring all requested values or variables to be correct. Detailed results are reported in Table~\ref{tab:ruler_1m_details}.

\begin{table}[t]
\centering
\footnotesize
\setlength{\tabcolsep}{6pt}
\caption{Detailed near-1M RULER-derived accuracy of 30B-A3B GSM (\%).}
\label{tab:ruler_1m_details}
\begin{tabular}{lrr}
\toprule
Task & Correct / Total & Accuracy (\%) \\
\midrule
4-key Retrieval & 32 / 32 & 100.00 \\
8-key Retrieval & 32 / 32 & 100.00 \\
2-hop Reasoning & 21 / 21 & 100.00 \\
4-hop Reasoning & 19 / 21 & 90.48 \\
8-hop Reasoning & 19 / 22 & 86.36 \\
\bottomrule
\end{tabular}
\end{table}

\subsection{Efficiency Measurement}
\label{app:efficiency_protocol}

All measurements use the 30B-A3B baseline and GSM checkpoints after 300 steps of 64K YaRN adaptation. We benchmark input lengths of 4,096, 16,384, 65,536, 131,072, and 262,144 tokens. Both models use matched text inputs and execution settings at each length. Inputs are drawn from PG19, with longer workloads constructed by cyclic repetition of corpus windows. The same adapted YaRN configuration is used throughout.

Training is benchmarked on 32 GPUs across four nodes (16 H800 and 16 H100), including forward computation, backward computation, gradient communication, and AdamW updates. Each configuration is warmed up for five steps and measured over ten updates. The duration of each step is the maximum elapsed time across all 32 ranks, and the reported step time is the median. Throughput is calculated from valid input tokens per update, excluding padding, divided by step time. Peak allocated GPU memory is the maximum CUDA allocation across ranks. Both models use identical optimizer, parallelism, and activation recomputation settings at each length.

Each training update processes 1,048,576 physical input positions, with 1,036,288, 1,045,504, 1,047,808, 1,048,192, and 1,048,384 valid tokens at the five respective lengths. At 4K and 16K, the data/context/expert parallelism configuration is 32/1/8, with microbatch--accumulation settings of 4$\times$2 and 1$\times$2. At 64K, 128K, and 256K, the parallelism configuration is 4/8/8, with settings of 1$\times$4, 1$\times$2, and 1$\times$1. Activation recomputation covers the first 12 layers at 4K--16K and all 28 backbone layers and the GSM decoder at 64K--256K.

In the complete training runs, the 4K joint text--image pretraining phase records mean step times of 3.163 seconds for GSM and 3.216 seconds for the baseline. During the 300-step 64K PG19 adaptation, the corresponding mean step times are 9.645 and 14.883 seconds. These measurements cover their respective full training workloads, whereas Figure~\ref{fig:inference_efficiency}(a) reports results from the controlled fixed-load benchmarks.

Inference is measured independently on H800 and H100 GPUs using bfloat16 weights, unquantized caches, tensor parallelism of one, and batch size one. Both models run sequentially on the same devices with matched CPU affinity and alternating execution order. Each model receives two warm-up requests and three measured requests per GPU, using three GPUs per configuration and four GPUs at 256K. Medians are reported over the nine or twelve measured requests for each GPU type and input length.

Each request consists of generation prefill followed by 64 sequential cached decoding steps. Prefill timing includes the full prompt computation required to construct historical KV and GSM shared states, evaluation of the first-token query, and full-vocabulary output projection. GSM processes the complete prompt in its encoder and evaluates only the final query position in its decoder during generation prefill. Decoding timing includes model computation, vocabulary projection, request-state updates, and runtime scheduling. A fixed continuation is supplied after each full-vocabulary logit computation to ensure matched decoding inputs. Prefill throughput is the number of input tokens divided by prefill time, and decoding throughput is 64 divided by continuation time. Both inference implementations use bounded history reads and CUDA Graph execution.

Request-cache storage is measured after prefill and before decoding by summing unique underlying allocations for the request state, including local windows, compressed history banks, pending entries, and GSM shared memory. Aliased storage is counted only once. Peak allocated GPU memory includes model weights, temporary workspaces, and CUDA Graph buffers and is measured across prefill and decoding. At 256K, GSM and the baseline use 0.295 and 0.491 GiB of request-cache storage, respectively, while their peak total allocations are 72.52 and 72.80 GiB.

\subsection{Shared Retrieval Ablation}
\label{app:retrieval_ablation}

The shared retrieval ablation compares two 0.5B GSM variants. Both use the same accumulated retrieval output and concatenated query inputs. One uses a separate aggregation indexer to select historical entries, while the other directly reuses the main attention Top-$K$ selections from the corresponding encoder layer. Both models undergo main training on approximately 10B tokens, 200 adaptation steps at 64K, and identical task adaptation with direct answers. Standard Avg.\ is the unweighted mean over ARC-Easy, PIQA, SciQ, BoolQ, and BLiMP. The first three tasks use \texttt{acc\_norm}, and the remaining two use \texttt{acc}.

To examine differences in the historical positions selected by the two retrieval mechanisms, we compare the aggregation indices with the corresponding main attention indices within the model that uses a separate indexer. Top-$K$ overlap is computed as the intersection size divided by $K$, considering only queries with more than $K$ valid candidates. Results are reported in Table~\ref{tab:topk_overlap}.

\begin{table}[t]
\centering
\small
\setlength{\tabcolsep}{7pt}
\caption{Overlap between aggregation and main attention indices (\%).}
\label{tab:topk_overlap}
\begin{tabular}{lrrr}
\toprule
Length & Stage 1 & Stage 2 & Stage 3 \\
\midrule
4K  & 43.008 & 51.139 & 55.604 \\
16K & 28.478 & 34.225 & 43.645 \\
32K & 23.445 & 28.565 & 39.441 \\
64K & 19.023 & 24.676 & 36.170 \\
\bottomrule
\end{tabular}
\end{table}

The overlap decreases consistently as context length increases, indicating that the separate indexer and main attention select increasingly different historical positions. Together with the task results in Table~\ref{tab:retrieval_ablation}, this shows that sharing the main attention Top-$K$ selections preserves comparable model performance despite substantially changing the candidate sets. A separate aggregation indexer is therefore not necessary for constructing the shared state.

\subsection{CoT Ablation}
\label{app:cot}

The CoT ablation starts from the same 1.5B GSM checkpoint adapted to 64K with YaRN and trains two variants: direct answer and short CoT. Both use the same 512 RULER retrieval training examples and are trained for 64 steps. Training lengths cycle through 4K, 8K, 16K, and 32K, while 64K is reserved exclusively for length extrapolation evaluation.

The direct-answer variant receives supervision only on the final answer. The short-CoT variant includes a brief retrieval and verification process in the target response and uses \texttt{Final answer:} to mark the final answer. Both variants use identical initial weights, input examples, and numbers of update steps, and evaluation considers only the final answers. Short CoT achieves 100\% accuracy at every tested length and on all three RULER tasks, as detailed in Tables~\ref{tab:ruler} and~\ref{tab:ruler_tasks}.

\end{document}